%% file: emnlp2018.tex
\documentclass[11pt,a4paper]{article}
\usepackage[hyperref]{emnlp2018}
\usepackage{times}
\usepackage{latexsym}
\usepackage{graphicx}
\usepackage{balance}
\usepackage{url}

\aclfinalcopy

\title{Identifying Domain Adjacent Instances for Semantic Parsers}

\author{  James Ferguson \\
  University of Washington \\
  {\normalsize {\tt jfferg@cs.washington.edu }} \\\And
  Janara Christensen \\
  Google Inc. \\
  \\\And
  Edward Li \\
  Google Inc \\
  {\normalsize {\tt \{jchristensen,gavinbelson,edgargip\}@google.com }}  \\\And
  Edgar Gonz\`alez \\
  Google Inc \\
   \\}

\date{}

\begin{document}
\maketitle

\input{abstract}

\input{introduction}

\input{problem}

\input{approach}

\input{evaluation}

\input{related_work}

\input{conclusion}

\bibliography{references}
\bibliographystyle{acl_natbib_nourl}

\end{document}

%% file: abstract.tex
\begin{abstract}
When the semantics of a sentence are not representable in a semantic parser's output schema, parsing will inevitably fail. Detection of these instances is commonly treated as an out-of-domain classification problem. However, there is also a more subtle scenario in which the test data is drawn from the same domain. In addition to formalizing this problem of \emph{domain-adjacency}, we present a comparison of various baselines that could be used to solve it. We also propose a new simple sentence representation that emphasizes words which are unexpected. This approach improves the performance of a downstream semantic parser run on in-domain and domain-adjacent instances.
\end{abstract}

%% file: introduction.tex
\section{Introduction}
\label{sec:introduction}

Semantic parsers map text to logical forms, which can then be used by downstream components to fulfill an action. Consider, for example, a system for booking air travel, in which a user provides natural language input, and a downstream subsystem is able to make or cancel flight reservations. Users of the system typically have a general understanding of its purpose, so the input will revolve around the correct topic of \textbf{air travel}. However, they are unlikely to know the limits of the system's functionality, and may provide inputs for which the expected action is beyond its capabilities, such as asking to change seats on a flight reservation. Because the logical schema is designed with fulfillment in mind, no logical form can capture the semantics of these sentences, making it impossible for the parser to generate a correct parse. Any output the parser generates will cause unintended actions to be executed downstream. For example, asking to change seats might be misparsed and executed as changing flights. Instead, the parser should identify that this input is beyond its scope so the condition can be handled.\footnote{In the final page of the paper, we suggest a few immediate downstream system behaviors when a domain-adjacent instance is identified, but others have investigated the related problem of teaching the system new behavior \cite{azaria_aaai16}.} In this paper, we formalize this pervasive problem, which we call \textit{domain-adjacent instance identification}.



\input{example_figure}


While this task is similar to that of identifying \emph{out-of-domain} input instances (e.g., \textbf{banking} with respect to \textbf{air travel}), it is much more subtle --- the instances come from roughly the same domain as the parser's training examples, and thus use very similar language. Domain adjacency is a property with respect to the parser's output schema, independent of the data used to train it.

In this paper, we formalize this task, and propose a simple approach for representing sentences in which words are weighted by how likely they are to differentiate between in-domain and domain-adjacent instances. Note that while this approach can also be applied to out-of-domain instances, in this paper we are interested in its performance on domain-adjacent instances. We describe an evaluation framework for this new task and, finally, evaluate our proposed method against a set of baselines, comparing performance on the domain-adjacent classification problem and a downstream semantic parsing task.  
 
%
%
%

%% file: example_figure.tex
\begin{figure}
\begin{center}
\small
\begin{tabular}{ |c|c|  }
 \hline
 \multicolumn{2}{|c|}{\textbf{Air Travel Domain}} \\
 \hline
 \multicolumn{2}{|c|}{Example In-Domain Predicates} \\
 \hline
 $buyTicket$ & Buy ticket LGA to SFO on 3/12    \\
 $flightStatus$ & What's the status of my SF flight?  \\
 $switchFlight$ & Change it to the 8am SFO flight    \\
 $cancelFlight$ & Cancel my flight to SFO \\
 $awardTravel$ & I want to fly to SFO with miles \\
 \hline
 \multicolumn{2}{|c|}{Example Domain-Adjacent Predicates} \\
 \hline
 $changeSeat$ & Change my seat to 23A \\
 $milesUpgrade$ & Upgrade my flight with my miles  \\
 $arrivalGate$ & Gate that my SFO flight arrives at\\
 $mileageStatus$ & What's my miles status \\
  \hline
  \multicolumn{2}{|c|}{Example Out-of-Domain Predicates} \\
 \hline
 $transferMoney$ & Transfer \$200 to checking   \\
 $addTimer$ & Add a timer for 3 minutes  \\
 $restaurantSearch$ & Thai restaurants in SF \\
 $scheduleMeeting$ & Set up a 9am meeting with Amy \\
 
 \hline
\end{tabular}
\end{center}
\caption{In this example, an \textbf{air travel} semantic parser is trained on data containing in-domain predicates. A test instance cannot be parsed correctly if it contains any domain-adjacent or out-of-domain predicates.}
%
%
%
%
\small
\label{fig:example}
\end{figure}

%% file: problem.tex
\section{Problem Setting}
\label{sec:problem}

A semantic parser can be seen as a function $\varphi$ that maps sentences $x$ in a natural language $\mathcal{L}$ to logical forms $y \in \mathcal{Y}$. Assuming the existence of an oracle parser $\hat{\varphi}$, the problem we propose in this paper is that of determining, for a given test instance $x$, whether it belongs to the \emph{domain} $\Phi$ of $\hat{\varphi}$, i.e., if its semantics can be encoded in the schema $\mathcal{Y}$.

In real-world usage, the input sentences $x$ will be generated by a human user, who associates the capabilities of the parser to a particular topic (e.g., \textbf{air travel}). Thus most of the $x \in \mathcal{L} \setminus \Phi$ will share topic with the $\hat{x} \in \Phi$. Because of the similarity between these $x$ and $\hat{x}$, we call this task \emph{identification of domain-adjacent instances}.

%% file: approach.tex
\section{Approach}
\label{sec:approach}
Our goal is to identify input instances whose semantics are not representable in the parser's output schema, and we assume only an in-domain dataset is available at training time. Our approach is based on determining similarity to these training instances. We split the task in two parts: 1) encode the sentences to a compact representation that preserves the needed information, and 2) given these representations, identify which sentences are so dissimilar that they are unlikely to be parseable with any schema that covers the training set.

\subsection{Sentence Representation}
\label{sec:sentence_representation}
Among recent work in distributional semantics, averaging the word vectors to represent a sentence \cite{wieting_iclr2016, adi_iclr2017} has proven to be a simple and robust approach. However, we have an intuition that words which are \emph{unexpected} in their context given the training data may be a strong signal that an instance is domain-adjacent. To incorporate this signal, we propose a weighted average, in which the weight corresponds to how unexpected the word is in its context. For example, given in-domain predicates from Figure~\ref{fig:example}, in the domain-adjacent sentence ``Upgrade my flight to SFO with my miles'', \emph{upgrade} should receive a much higher weight than  \emph{flight} or \emph{SFO}.

Our weighting scheme is as follows: We use the cosine distance between the expected ($\bar{v}_i$) and the actual ($\hat{v}_i$) domain-specific word embedding at a given position ($i$) in a sentence to compute its weight: $w_i = 1 - cos(\bar{v}_i, \hat{v}_i)$. The expected word embedding is computed using the context embeddings, $\bar{v}_i = \sum_{j=i-c, j\neq i}^{i+c} \hat{v}_j$, where $\hat{v}_j$ is a domain-specific word embedding, in a window of size $c$ around position $i$. Intuitively, $w_i$ represents how surprising the word is in the context.


%
%
%
%
%

Since our training set is too small to directly learn domain-specific embeddings, we learn a mapping from general pre-trained embeddings. We train a continuous bag-of-words model \cite{mikolov_iclr2013} in which we pass pre-trained embeddings ($v_i$) instead of 1-hot vectors, as input to the embedding layer. The layer thus learns a mapping from pre-trained to domain-specific embeddings ($\hat{v}_i$). We use this mapping to compute new embeddings for words that are missing from the training set. Only words that do not have pre-trained embeddings are ignored.

Finally, for a sentence with $n$ words, we take the weighted average of the pre-trained embeddings of the words in the sentence, using the weights from above: $S = \left( \sum_{i=1}^{n} w_i v_i \right) / \left( \sum_{i=1}^{n} w_i \right)$.
%
%
%
%

This approach assigns high weight to words that differ significantly from what is expected based on the training data. By combining these weights with the pre-trained word embeddings, we allow the model to incorporate external information, improving generalization beyond the training set.
%
%
%
%

\subsection{Domain-Adjacent Model}
\label{sec:ood_model}
A number of techniques can be applied to predict whether a sentence is domain-adjacent from its continuous representation. Of the methods we tried, we found k-nearest neighbors \citep{angiulli2002fast} to perform best: to classify a sentence, we calculate the average cosine distance between its embedding and its $k$ nearest neighbors in the training data, and label it domain-adjacent if this value is greater than some threshold. This simpler model relies more heavily on the external information brought in by pre-trained word embeddings, while more complex models seem to overfit to the training data. 
%
%

%% file: evaluation.tex
\section{Evaluation}
%
%
\label{sec:evaluation}
In this section, we introduce an evaluation framework for this new task. We consider training and test sets from a single domain, with only the latter containing domain-adjacent instances. 
Test instances are classified individually, and we measure performance on in-domain/domain-adjacent classification and semantic parsing. 

\subsection{Dataset and Semantic Parser}
\label{sec:experiment_setup}
\input{predicates}
We simulate this setting by adapting the \textsc{Over\-night} dataset \cite{wang_acl2015}. 
This dataset is composed of queries drawn from eight domains, each having a set of seven to eighteen distinct semantic predicates. Queries consist of a crowd-sourced textual sentence and its corresponding logical form containing one or more of these domain-specific semantic predicates.

For each domain, we select a set of predicates to exclude from the logical schema (see Table \ref{table:unsupported_predicates}), and remove all instances containing these predicates from the training set (since they are now domain-adjacent). We then train a domain-adjacent model and semantic parser on the remaining training data and attempt to identify the domain-adjacent examples in the test data. We use the train/test splits from \citet{wang_acl2015}. 
In all experiments, we use the SEMPRE parser \cite{berant_emnlp2013}.

\subsection{Baselines}
\label{sec:experiment_baselines}

Because this is a novel task, and results are not comparable to previous work, we report results from a  variety of baseline systems.
The first baseline, \textsc{Confidence}, identifies instances as domain-adjacent if the semantic parser's confidence in its predictions is below some threshold. 

The remaining baselines follow the two-part approach from Section \ref{sec:approach}.
\textsc{Autoencoder} is inspi\-red by \citet{ryu_pattern2017}'s work on identifying \emph{out-of-domain} examples. For the sentence representation, this method uses a bi-LSTM with self-attention, trained to predict the semantic predicates, and concatenates the final hidden state from each direction as the sentence representation. An autoencoder is used as the domain-adjacent classifier. 

The remaining methods use the nearest neighbor model discussed in Section~\ref{sec:ood_model}. For sentence representations, we include baselines drawn from different neural approaches. In \textsc{CBOW}, we simply average the pre-trained word embeddings in the sentence. In \textsc{CNN}, we train a two-layer CNN with a final softmax layer to predict the semantic predicates for a sentence. We concatenate the mean pooling of each layer as the sentence representation. In \textsc{LSTM}, we use the same sentence representation as in \textsc{Autoencoder}, with the nearest neighbor domain-adjacent model. Finally, \textsc{Surprise} is the approach presented in Section~\ref{sec:sentence_representation}.

\subsection{Direct Evaluation}
\label{sec:experiment_direct_evaluation}
\input{direct_results_sentrep}

We first directly evaluate the identification of domain-adjacent instances: Table \ref{table:direct_eval_sentencerep} reports the area under a receiver operating characteristic curve (AUC) for the considered models \cite{fawcett_pr2006}.
\textsc{Surprise} generally performs the best on this evaluation; and, in general, the simpler models tend to perform better, suggesting that more complex approaches tune too much to the training data.

Qualitatively, for domains where the \textsc{Surprise} model performs better, it places higher weight on words we would consider important for distinguishing domain-adjacent sentences. 
For example in ``show me recipes with longer preparation times than rice pudding'' from \textbf{Recipes}, ``longer'' and ``preparation'' have the highest weights. 
In \textbf{Social}, there are two in-domain predicates (\emph{employmentStartDate} and \emph{educationEndDate}) which use very similar wording to those that are domain-adjacent, making it difficult to isolate surprising words. The weights in this domain seem to instead emphasize unusual wordings such as ``soonest'' in ``employees with the soonest finish date''.

%
%
%
%
%

\subsubsection{Ablation Analysis}
\label{sec:experiment_direct_evaluation_ablation}
\input{ablation_results}

In order to determine the contribution of each one of the components of \textsc{Surprise}, we performed an ablation analysis comparing the following modifications of the method: \textsc{CBOW}, as described above, using an unweighted average of pre-trained embeddings; \textsc{Frequency}, using a weighted average of pre-trained embeddings, with weights based on inverse document frequency in the training set; \textsc{PreTrained}, using the surprise schema but with weights determined using pre-trained embeddings; and the full \textsc{Suprise} as presented above. Each approach adds one component (weighting, surprise-based weights, and domain-specific embeddings) with respect to the previous one.

The results of the experiment are shown in Table~\ref{table:ablation_results}. We can see that \textsc{Frequency} performs slightly worse than \textsc{CBOW} and \textsc{PreTrained} performs even worse than that. We can conclude that the combination of the weighting schema and the tuned vectors is what makes \textsc{Suprise} effective.

\subsection{Downstream Task Evaluation}
\label{sec:experiment_downstream_task_evaluation}
\input{downstream_results_acc}
We next evaluate how including the domain-adjacent predictions affects the performance of a semantic parser. 
In a real setting, when the semantic parser is presented with domain-adjacent input that is beyond its scope, the correct behavior is to label it as such so that it can be handled properly by downstream components. To simulate this behavior, we set the gold parse for domain-adjacent instances to be an empty parse, and automatically assign an empty parse to any instance that is identified as domain-adjacent. We report accuracy of the semantic parser with 20\% domain-adjacent test data. We include two additional models: \textsc{NoFilter}, in which nothing is labeled domain-adjacent, and \textsc{Oracle}, in which all the domain-adjacent instances are correctly identified.
For each baseline requiring a threshold, we set it such that 3\% of the instances in the dev set would be marked as domain-adjacent (intuitively, this represents the error-tolerance of the system).

Table \ref{table:downstream_eval} shows the results for this experiment. In general, the relative performance is similar to that in the direct evaluation (e.g. \textsc{Surprise} tends to do well on most domains, but performs poorly on \textsc{Basketball} and \textsc{Social} in particular).
However, in this evaluation, misclassifying an instance as domain-adjacent if the semantic parser would have accurately parsed it is worse than misclassifying the instance if the semantic parser could not have accurately parsed it. For example, in \textsc{Social} we can thus infer that \textsc{Surprise} is marking some instances as domain-adjacent that would otherwise be accurately parsed as the performance there is actually worse than for \textsc{NoFilter}.


%
%
%
%

%% file: predicates.tex
\begin{table}
\begin{center}
\small
\begin{tabular}{|c|c|}
\hline
Basketball & \emph{numGamesPlayed} \\ \hline
Blocks  & \emph{length} \\ \hline
Calendar & \emph{startTime} \\ \hline
Housing & \emph{size} \\ \hline
Publications & \emph{venue} \\ \hline
Recipes & \emph{preparationTime} \\ \hline
Restaurants & \emph{starRating} \\ \hline
Social  & \emph{educationStartDate}, \\
& \emph{employmentEndDate} \\
\hline
\end{tabular}
\end{center}
\caption{Predicates excluded from training and considered domain-adjacent. Domains have 5-20 predicates.}

\label{table:unsupported_predicates}
\end{table}

%% file: direct_results_sentrep.tex
\begin{table*}
\begin{center}
\resizebox{\textwidth}{!}{
\begin{tabular}{|c|c|c|c|c|c|c|c|c|}
\hline
& Basketball & Blocks & Calendar & Housing & Publications  & Recipes & Restaurants & Social \\
\hline
\textsc{Autoencoder} & 0.801 & 0.479 & 0.766 & 0.781 & 0.874 & 0.722 & 0.581 & 0.774\\
\hline
\textsc{Confidence}& 0.660 & 0.738 & 0.697 & 0.648 & 0.631 & 0.651 & 0.730 & 0.573  \\
\hline
\textsc{CBOW}      & 0.743 & 0.782 & 0.662 & 0.910 & 0.884 & 0.670 & 0.911 & 0.675 \\
\textsc{CNN}       & \textbf{0.916} & 0.654 & \textbf{0.862} & 0.792 & 0.908 & 0.505 & 0.840 & \textbf{0.813} \\
\textsc{LSTM} & 0.826 & 0.571 & 0.741 & 0.912 & 0.827 & 0.487 & 0.593 & 0.754 \\
\textsc{Surprise}  & 0.755 & \textbf{0.827} & 0.817 & \textbf{0.933} & \textbf{0.978} & \textbf{0.758} & \textbf{0.941} & 0.545 \\
\hline
\end{tabular}
}
\end{center}
\caption{AUC for domain-adjacent instance identification, using \textsc{kNN} as the domain-adjacent model.}
\label{table:direct_eval_sentencerep}
\end{table*}

%% file: ablation_results.tex
\begin{table*}[t]
\begin{center}
\resizebox{\textwidth}{!}{
\begin{tabular}{|c|c|c|c|c|c|c|c|c|}
\hline
& Basketball & Blocks & Calendar & Housing & Publications  & Recipes & Restaurants & Social \\
\hline
\textsc{CBOW} &  0.743 &  0.782 &  0.662 &  0.910 &  0.884 &  0.670 &  0.911 &  \textbf{0.675} \\
\hline
\textsc{Frequency}  &  0.656 &  0.703 &  0.771 &  0.884 &  0.887 &  0.667 &  0.834 &  0.591 \\
\hline
\textsc{PreTrained} & 0.612 & 0.636 & 0.512 & 0.819 & 0.842 & 0.526 & 0.858 & 0.538 \\
\hline
\textsc{Surprise}   &  \textbf{0.755} &  \textbf{0.827} &  \textbf{0.817} &  \textbf{0.933} &  \textbf{0.978} &  \textbf{0.758} &  \textbf{0.941} &  0.545 \\
\hline
\end{tabular}
}
\end{center}
\caption{AUC for domain-adjacent instance identification, using ablated versions of \textsc{Surprise} with \textsc{kNN}.}
\label{table:ablation_results}
\end{table*}

%% file: downstream_results_acc.tex
\begin{table*}
\begin{center}
\resizebox{\textwidth}{!}{
\begin{tabular}{|c|c|c|c|c|c|c|c|c|}
\hline
& Basketball & Blocks & Calendar & Housing & Publications  & Recipes & Restaurants & Social \\
\hline
\textsc{NoFilter} & 0.358 & 0.294 & 0.617 & 0.461 & 0.511 & 0.570 & 0.626 & 0.355\\
\textsc{Oracle} & 0.558 & 0.494 & 0.817 & 0.661 & 0.711 & 0.770 & 0.826 & 0.555 \\
\hline
\textsc{Autoencoder} & 0.413 & 0.268 & 0.581 & 0.447 & 0.463 & 0.530 & 0.543 & 0.417\\
\hline
\textsc{Confidence} & 0.389 & 0.306 & 0.644 & 0.472 & 0.525 & 0.568 & 0.665 & 0.360\\
\hline
\textsc{CBOW} & 0.344 & 0.295 & 0.634 & 0.515 & 0.621 & \textbf{0.575} & 0.722 & 0.358\\
\textsc{CNN} & \textbf{0.452} & 0.324 & 0.674 & 0.488 & 0.573 & 0.570 & 0.605 & \textbf{0.446}\\
\textsc{LSTM} & 0.385 & 0.314 & 0.622 & 0.495 & 0.581 & 0.547 & 0.612 & 0.363\\
\textsc{Surprise} & 0.356 & \textbf{0.371} & \textbf{0.679} & \textbf{0.570} & \textbf{0.668} & 0.554 & \textbf{0.764} & 0.345\\
\hline
\end{tabular}
}
\end{center}
\caption{Accuracy for a semantic parser evaluated on a test set in which 20\% is domain adjacent.}
\label{table:downstream_eval}
\end{table*}

%% file: related_work.tex
\section{Related Work}
\label{sec:related_work}
%
%
%


Domain-adjacency identification is a new task, but relatively little effort has been devoted 
to even the related task of identifying out-of-domain instances (i.e., from completely separate domains) for semantic parsers. \citet{hakkani-tur_interspeech2015} approached the problem by clustering sentences based on shared subgraphs in their general semantic parses; \citet{ryu_pattern2017} classify sentences with autoencoder reconstruction error.


Prior distributional semantics work to create compact sentential representations generated specific embeddings for downstream tasks \cite{kalchbrenner_acl2014, kim_emnlp2014, socher_emnlp2013}. Recently, work has focused on domain-independent embeddings, learned without downstream task supervision. \citet{kiros_nips2015}, \citet{hill_naacl2016}, and \citet{kenter_acl2016} learn representations by predicting the surrounding sentences. \citet{wieting_iclr2016} use paraphrases as\linebreak{}supervision. \citet{mu_acl2017} represent sentences\linebreak{} by the low-rank subspace spanned by the embeddings of the words in them; \citet{arora_iclr2017} use a weighted average of word embeddings, with their projection onto the first principal component across all sentences in the corpus removed.

Another relatively sparse area of related work is handling the domain-adjacent instances once they have been identified. The simplest thing to do is to return a generic error. For user-facing applications, one such message might state that the system can't handle that specific query. \citet{azaria_aaai16} approach this problem by having the user break down the domain-adjacent instance into a sequence of simpler textual instructions and then attempting to map those to known logical forms.

%% file: conclusion.tex
\section{Conclusion}
%
\label{sec:conclusion}


Identifying domain-adjacent instances is a practical issue that can improve downstream semantic parsing precision, and thus provide a smoother and more reliable user experience. 
In this paper, we formalize this task, and introduce a novel sentence embedding approach which outperforms baselines. Future work includes exploring alternative ways of incorporating information outside of the given training set and experimenting with various combinations of semantic parsers and upstream domain-adjacency models. Another area of future research is how the underlying system should recover when domain-adjacent instances are detected.